\def\year{2021}\relax
\documentclass[letterpaper]{article} 
\usepackage{aaai21}  
\usepackage{times}  
\usepackage{helvet} 
\usepackage{courier}  
\usepackage[hyphens]{url}  
\usepackage{graphicx} 
\usepackage{natbib}  
\usepackage{caption} 
\usepackage{amssymb}
\usepackage{multirow}
\usepackage{siunitx}
\usepackage[switch]{lineno}
\title{Keyword-Guided Neural Conversational Model}
\author {
    Peixiang Zhong,\textsuperscript{\rm 1,2}
    Yong Liu, \textsuperscript{\rm 1,2}
    Hao Wang, \textsuperscript{\rm 3}\thanks{Corresponding author}
    Chunyan Miao \textsuperscript{\rm 1,2}\footnotemark[1] \\
}
\affiliations {
    \textsuperscript{\rm 1} Alibaba-NTU Singapore Joint Research Institute, Nanyang Technological University (NTU), Singapore\\
    \textsuperscript{\rm 2} Joint NTU-UBC Research Centre of Excellence in Active Living for the Elderly, NTU, Singapore\\
    \textsuperscript{\rm 3} Alibaba Group, China \\
    \{peixiang001, stephenliu, ascymiao\}@ntu.edu.sg, cashenry@126.com
}
\begin{document}

\maketitle

\begin{abstract}
We study the problem of imposing conversational goals/keywords on open-domain conversational agents, where the agent is required to lead the conversation to a target keyword smoothly and fast. Solving this problem enables the application of conversational agents in many real-world scenarios, e.g., recommendation and psychotherapy. The dominant paradigm for tackling this problem is to 1) train a next-turn keyword classifier, and 2) train a keyword-augmented response retrieval model. However, existing approaches in this paradigm have two limitations: 1) the training and evaluation datasets for next-turn keyword classification are directly extracted from conversations without human annotations, thus, they are noisy and have low correlation with human judgements, and 2) during keyword transition, the agents solely rely on the similarities between word embeddings to move closer to the target keyword, which may not reflect how humans converse. In this paper, we assume that human conversations are grounded on commonsense and propose a keyword-guided neural conversational model that can leverage external commonsense knowledge graphs (CKG) for both keyword transition and response retrieval. Automatic evaluations suggest that commonsense improves the performance of both next-turn keyword prediction and keyword-augmented response retrieval. In addition, both self-play and human evaluations show that our model produces responses with smoother keyword transition and reaches the target keyword faster than competitive baselines.
\end{abstract}

\section{Introduction}
\label{sec: introduction}
Building a human-like open-domain conversational agent (CA) has been one of the milestones in artificial intelligence (AI). Early conversational agents are primarily based on rules \cite{weizenbaum1966eliza, colby1971artificial}, e.g., Eliza \cite{weizenbaum1966eliza}, the first CA developed in 60's, simulates a Rogerian psychotherapist based on hand-crafted pattern matching rules. In recent years, with the advancement of data-driven neural networks, neural open-domain conversational models are becoming dominant \cite{vinyals2015neural, lowe2015ubuntu, gao2018neural}.

Recent efforts in open-domain neural conversational models are primarily aiming to improve the response diversity \cite{li2016simple, zhang2018generating} and endowing responses with knowledge \cite{zhou2018commonsense, dinan2018wizard}, personality \cite{li2016persona, zhang2018personalizing}, emotion \cite{zhou2018emotional, zhong2019affect} and empathy \cite{rashkin2019towards, zhong2020towards}.
\begin{figure}[!t]
\centering
\includegraphics[width=1\linewidth]{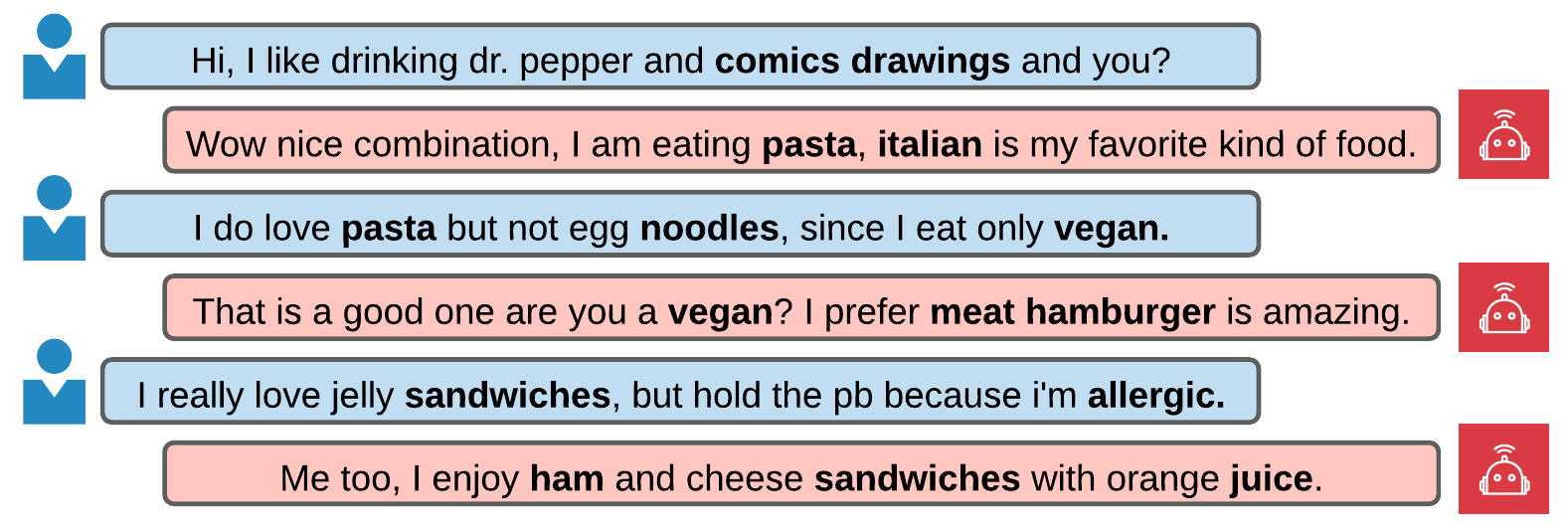}
\caption{Illustration of keyword-guided conversations from self-play simulations. Keywords are highlighted in bold. Given a random starting keyword ``comics", the agent (red) leads the conversation to the target keyword ``juice" smoothly and fast.}
\label{fig: example conversation}
\end{figure}
All the efforts mentioned above are focusing on models that passively respond to user messages. However, in many real-world scenarios, e.g., conversational recommendation, psychotherapy and education, conversational agents are required to actively lead the conversation by smoothly changing the conversation topic to a designated one. For example, during a casual conversation, the agent may actively lead the user to a specific product or service that the agent wants to introduce and recommend.

In this paper, we follow the line of research in \cite{tang2019target, qin2020dynamic} and study the problem of imposing conversational goals/keywords on open-domain conversational agents, where the agent is required to lead the conversation to a target keyword smoothly and fast. As illustrated in Figure \ref{fig: example conversation}, given a target keyword ``juice" and a random starting keyword ``comics", the agent is required to converse with the user in multiple exchanges and lead the conversation to ``juice". The challenge of this problem lies in how to balance the tradeoff between maximizing keyword transition smoothness and minimizing the number of turns taken to reach the target. On the one hand, passively responding to the user solely based on the conversation context would achieve high smoothness but may take many turns to reach the target, but on the other hand, directly jumping to the target word by ignoring the conversation context would minimize the number of turns but produce non-smooth keyword transitions.

\citet{tang2019target} proposed to break down the problem into two sub-problems: next-turn keyword selection and keyword-augmented response retrieval. \citet{tang2019target} proposed a next-turn keyword predictor and a rule-based keyword selection strategy to solve the first sub-problem, allowing the agent to know what is the next keyword to talk about given the conversation history and the target keyword. In addition, \citet{tang2019target} proposed a keyword-augmented response retrieval model to solve the second sub-problem, allowing the agent to produce a response that is relevant to the selected keyword. 
\begin{figure}[!t]
\centering
\includegraphics[width=0.75\linewidth]{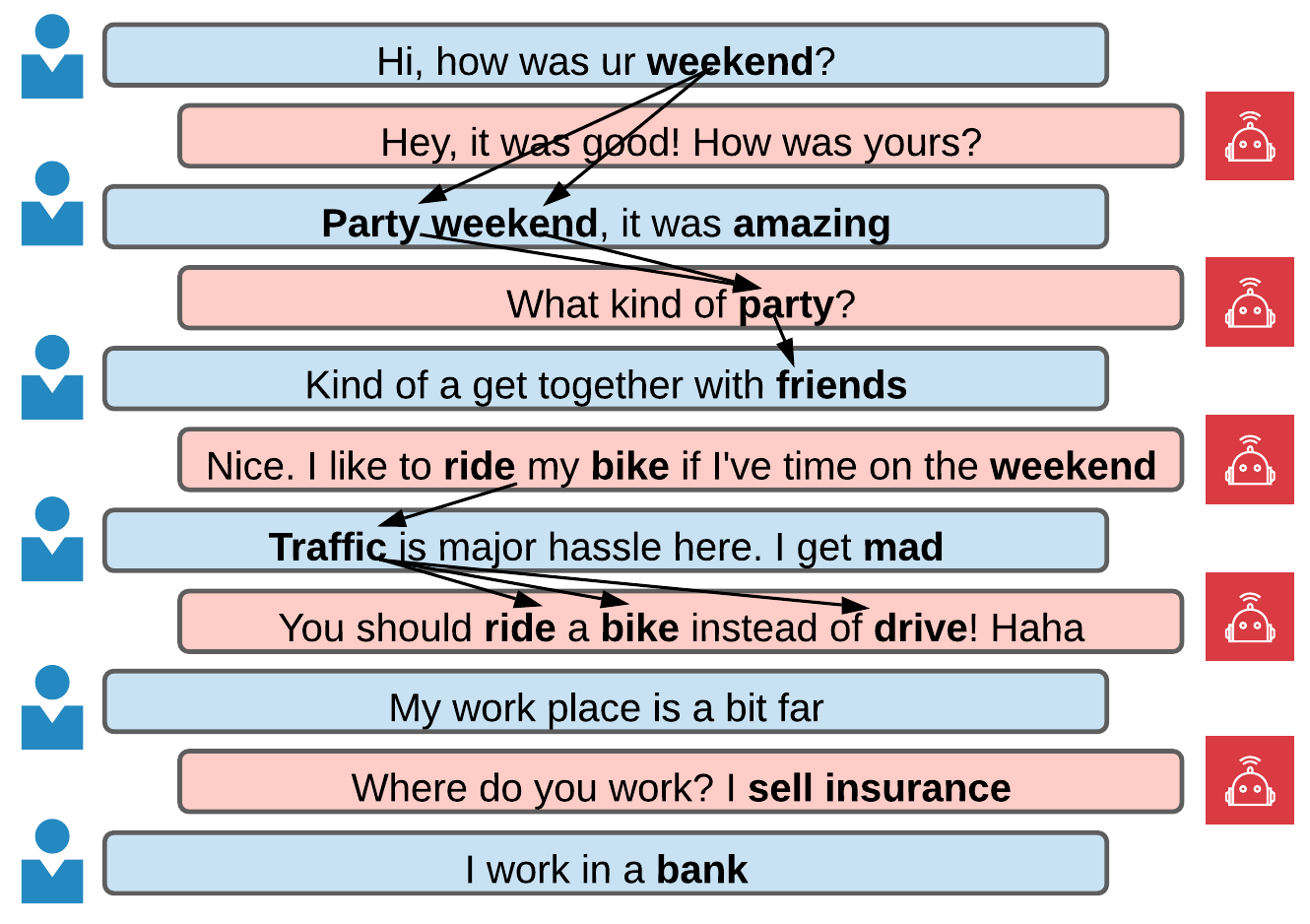}
\caption{Illustration of keyword (in bold) transitions in a sample conversation from ConvAI2 \cite{zhang2018personalizing}. Transitions indicated by arrows are considered relevant. The rest keyword transitions, e.g., friends $\,\to\,$ ride, are irrelevant (but used in the training and evaluation datasets of existing studies).}
\label{fig: topic transition}
\end{figure}

However, there are two major limitations in existing studies \cite{tang2019target, qin2020dynamic}. First, the training and evaluation datasets for next-turn keyword prediction are directly extracted from conversations without human annotations, thus, the majority of the ground-truth keyword transitions are noisy and have low correlations with human judgements. As illustrated in Figure \ref{fig: topic transition}, only a few keyword transitions in a conversation are considered relevant. In fact, in our human annotation studies of over 600 keyword transitions, we found that around 70\% of keyword transitions in the next-turn keyword prediction datasets are rated as not relevant, which renders the trained next-turn keyword predictor in existing studies less reliable. 
Second, the rule-based keyword selection strategy primarily leverages the cosine similarity between word embeddings to select keywords that are closer to the target keyword. Word embeddings are trained based on the distributional hypothesis that words that have similar contexts have similar meanings, which may not reflect how humans relate words in conversational turn-taking.

In this paper, we assume that human conversations are grounded on commonsense and propose a keyword-guided neural conversational model that can leverage external commonsense knowledge graphs (CKG) for both next-turn keyword selection and keyword-augmented response retrieval. Humans rely on commonsense to reason, and commonsense reasoning plays an important role in the cognitive process of conversational turn-taking \cite{schegloff1991conversation, stocky2004commonsense, lieberman2004beating}. Relying on a CKG for keyword transition would allow the agent to select a more target-related keyword for the next-turn. 
Moreover, we leverage commonsense triplets from the CKG using Graph Neural Networks (GNN) for both next-turn keyword prediction and keyword-augmented response retrieval to achieve more accurate predictions. 

In summary, our contributions are as follows:
\begin{itemize}
    \item We identify two limitations of existing studies in next-turn keyword selection: 1) noisy training and evaluation datasets, and 2) unreliable keyword transition based on the similarity between word embeddings.
    \item For the first time in this task, we propose to use CKG for keyword transition and propose two GNN-based models to incorporate commonsense knowledge for next-turn keyword prediction and keyword-augmented response retrieval, respectively.
    \item We propose a large-scale open-domain conversation dataset for this task, obtained from Reddit. The linguistic patterns in Reddit are far more diverse than the ConvAI2 \cite{zhang2018personalizing} used in existing studies, which are collected from only hundreds of crowd-workers. 
    \item We conduct extensive experiments and the results show that grounding keyword transitions on CKG improves overall conversation smoothness and allows the agent to reach the target faster. In addition, leveraging commonsense triplets substantially improves the performance of both next-turn keyword prediction and keyword-augmented response retrieval. Finally, self-play and human evaluations show that our model produces smoother responses and reaches the target keyword faster than competitive baselines.
\end{itemize}
\section{Related Work}
In recent years, several studies proposed to build conversational agents that can actively lead a conversation to a designated target keyword/goal \cite{tang2019target, wu2019proactive}. Our work follows the task definition in \cite{tang2019target}, which has been discussed in Introduction. Very recently, \citet{qin2020dynamic} improved \cite{tang2019target} in 1) next-turn keyword prediction by only considering keyword transitions that are present in the training dataset and 2) keyword-augmented response retrieval by constraining that the selected response must contain the predicted keyword or a keyword closer to the target keyword. As a result, \citet{qin2020dynamic} obtained the state-of-the-art performance on this task in terms of task success rate and transition smoothness.

Another line of research \cite{wu2019proactive} focused on the specific movie domain and proposed to use factoid knowledge graph to proactively lead the conversation from a random entity to a given entity. Our work differs from \cite{wu2019proactive} in that 1) we focus on open-domain conversations whereas they focus on movie domain; 2) we leverage commonsense knowledge graph for keyword transitions whereas they leverage factoid knowledge graph for entity transitions\footnote{In our work, a keyword can be a named entity, e.g., AAAI2021, or a generic content word, e.g., conference.}; and 3) we allow the target to be any arbitrary keyword whereas they constrain the target to be at most two-hop away from the starting entity. Following the line of research in \cite{wu2019proactive}, \citet{xu2020knowledge} proposed to use hierarchical reinforcement learning (HRL) to incorporate factoid knowledge graph for high-level topic selection and low-level in-depth topic-related conversation. \citet{xu2020conversational} proposed a framework to represent prior information as a conversation graph (CG) and leverage policy learning to incorporate the CG into conversation generation.

Commonsense has been studied extensively in recent neural conversational models \cite{young2018augmenting, zhou2018commonsense, zhang2020grounded, zhong2020care}. 
\citet{zhou2018commonsense} proposed graph attentions to statically incorporate one-hop knowledge triplets into conversation understanding and dynamically generate knowledge-aware responses. Recently, \citet{zhang2020grounded} extended \cite{zhou2018commonsense} to multi-hop knowledge triplets by proposing an attention mechanism to incorporate outer triplets and a GNN model to aggregate central triplets. Different from existing studies that leverage commonsense to improve the diversity and informativeness of responses, we incorporate commonsense into our approach for more reasonable keyword transition and more accurate response retrieval.

\section{Our Approach}
\label{sec: our approach}
In this section, we first introduce our task definition, and then describe the CKG used in our paper, and finally propose the \textbf{C}ommonsense-aware \textbf{K}eyword-guided neural \textbf{C}onversational model (CKC).

\subsection{Task Definition}
\label{sec: task definition}
Given a conversation history of $n$ utterances: $x_{1:n} = x_1, ..., x_n$, we denote the sequence of keywords for $x_i$ as $k_i$, and the response to $x_{1:n}$ as $y$. 

Briefly, given a target keyword $t$ and a random initial utterance $x_1$ with its keywords $k_1$, the task of the agent is to chat with the user and lead the conversation to the target keyword smoothly and fast. The target is only presented to the agent and unknown to the user. We consider the target is achieved when an utterance (either by the user or by the agent) mentions the target keyword\footnote{This is different from \cite{tang2019target} where mentioning a synonym of the target can be considered as success because we found that synonyms are unreliable to measure the task success.}.

We break down the task into two sub-problems: next-turn keyword selection and keyword-augmented response retrieval. We propose a CKG-aware next-turn keyword predictor and a CKG-guided keyword transition strategy to solve the first sub-problem. We then propose a CKG-aware keyword-augmented response retrieval to solve the second sub-problem.

\subsection{Commonsense Knowledge Graph (CKG)}
\label{sec: ckg}
In this paper, we use ConceptNet \cite{speer2017conceptnet} as our CKG. ConceptNet is a large-scale multilingual semantic graph that describes general human knowledge in natural language. Each node/concept on ConceptNet can be a single word, e.g., ``food" or a multi-word expression, e.g., ``having\_lunch". The edges on ConceptNet represent the semantic relations between nodes and have weights suggesting the confidence score, e.g., $\langle$\textit{having\_lunch}, \textit{HasPrerequisite}, \textit{food}$\rangle$ with a weight of 2.83. The majority of edge weights are in [0, 10]. We only include triplets that satisfy the following requirements into our CKG: 1) the edge weight is at least 1, 2) at least one node is in our keyword vocabulary\footnote{The keyword vocabulary is a subset of our word vocabulary containing frequent content words.}, and 3) the other node is in our word vocabulary\footnote{For a multi-word expression, we require that each single word to be in our word vocabulary.}.

\subsection{CKG-Aware Next-Turn Keyword Prediction}
\label{sec: keyword prediction}
\begin{figure}[!t]
\centering
\includegraphics[width=1\linewidth]{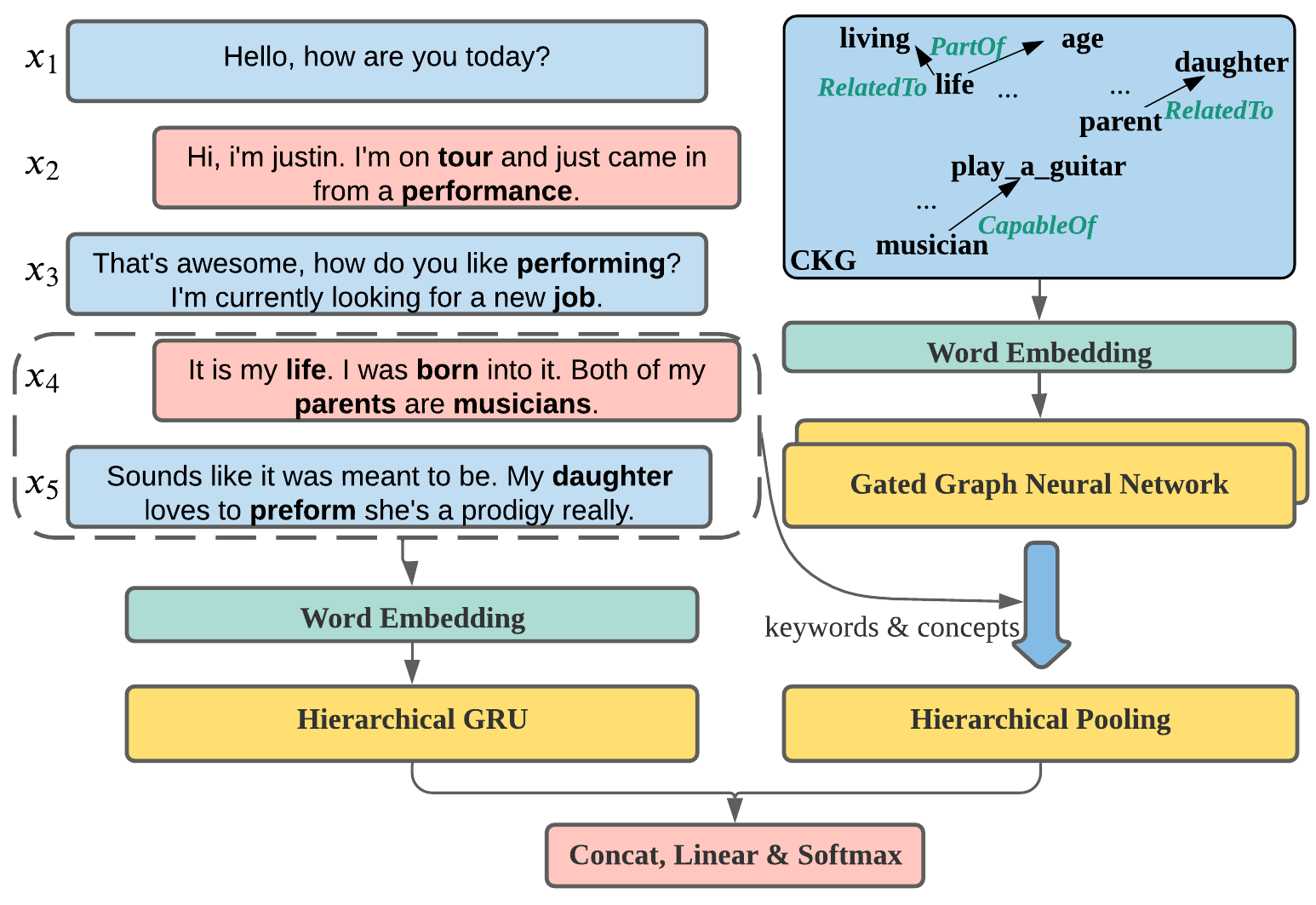}
\caption{Illustration of our proposed CKG-aware next-turn keyword prediction. We only use the most recent two utterances and their concepts and keywords as input. Words in bold denote keywords. Concepts are words or multi-word expressions extracted from utterances based on the CKG vocabulary.}
\label{fig: keyword prediction}
\end{figure}
\begin{figure*}[!t]
\centering
\includegraphics[width=0.9\linewidth]{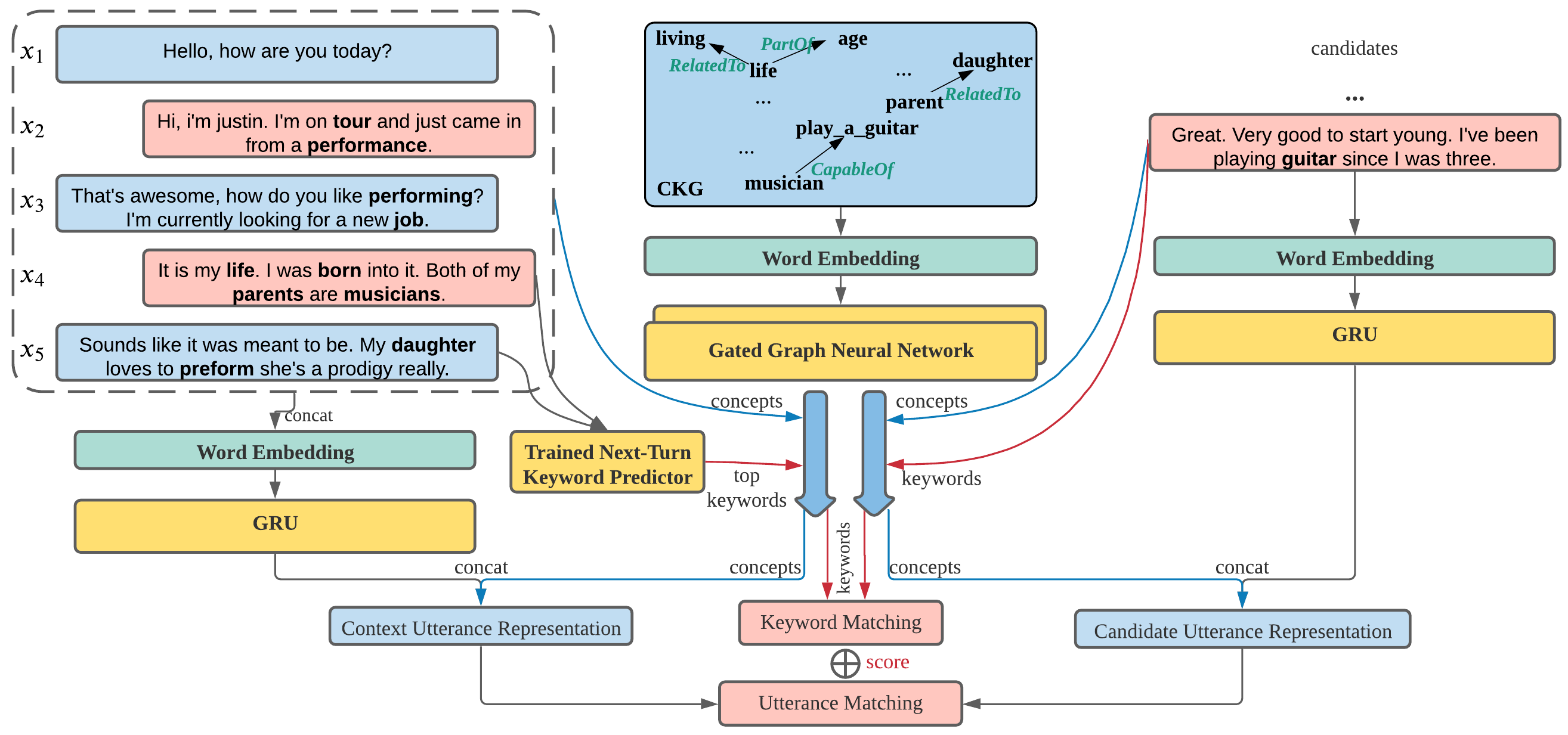}
\caption{Illustration of our proposed CKG-aware response retrieval model.}
\label{fig: response retrieval}
\end{figure*}
Given a history of $n$ utterances $x_{1:n}$ and $n$ sequences of keywords $k_{1:n}$, we propose a model that can predict the next-turn keywords $k_{n+1}$. Note that $k_{n+1}$ can include multiple keywords, hence this is a multi-label classification problem.

One major limitation of existing studies is that the training and evaluation datasets for next-turn keyword prediction are noisy, as discussed in Introduction. In this paper, we assume that human conversations are grounded on commonsense and leverage commonsense to 1) clean the training and evaluation datasets; and 2) propose a CKG-aware model for more accurate next-turn keyword prediction. 

Specifically, for each example in both training and evaluation datasets, we remove next-turn keywords that are not in the immediate neighborhood of historical keywords. During model prediction in both training and evaluation, we also only output keywords that are in the immediate neighborhood of input keywords. In other words, our model only outputs CKG-grounded keyword predictions.

We then propose a CKG-aware model that takes as input $x_{n-1}, x_{n}, k_{n-1}, k_{n}$ and the CKG, and output $k_{n+1}$. Note that existing studies only use $k_{n-1}, k_{n}$ and GRU \cite{cho2014learning} to predict $k_{n+1}$ \cite{tang2019target, qin2020dynamic}. Using longer context information does not improve performance in our experiments. An illustration of our model is presented in Figure \ref{fig: keyword prediction}. 

\subsubsection{Utterance Representation}
We obtain the utterance representation $\mathbf{x} \in \mathbb{R}^{d_1}$ from the contextual utterances $x_{n-1}$ and $x_{n}$ using a hierarchical GRU (HGRU) encoder, where $d$ denotes the final hidden state size of HGRU. 
\subsubsection{CKG Graph Representation}
We obtain a CKG graph representation $\mathbf{G} \in \mathbb{R}^{N\times d_2}$ using a Gated Graph Neural Network (GGNN) \cite{li2016gated}, where $N$ denotes the number of nodes in the CKG and $d_2$ denotes the hidden size of GGNN. For each node on the CKG, the convolution operation in GGNN first computes a parameterized weighted average of neighboring node representations and then updates its own representation using a GRU. The nodes in CKG are represented via word embeddings. Multi-word nodes are represented via averaged word embeddings. The CKG representation is learned jointly with the next-turn keyword prediction and the gradients on the CKG are directly back-propagated to the word embeddings. Both utterances and CKG share the same word embedding layer, which can effectively reduce the number of model parameters and enable knowledge transfer on word embeddings.
\subsubsection{Keyword and Concept Representation}
We extract the keyword and concept representations $\mathbf{K} \in \mathbb{R}^{N_k \times d_2}$ and $\mathbf{C} \in \mathbb{R}^{N_c \times d_2}$ from $\mathbf{G}$, respectively, where $N_k = |k_{n-1}| + |k_{n}|$ and $N_c$ denote the number of concepts in $x_{n-1}$ and $x_{n}$. Concepts are extracted from utterances via string matching with the CKG. We then apply hierarchical pooling where we first use mean pooling to aggregate $\textbf{K}$ and $\textbf{C}$ and obtain $\textbf{k} \in \mathbb{R}^{d_2}$ and $\textbf{c} \in \mathbb{R}^{d_2}$, respectively, and then apply max pooling to combine $\textbf{k}$ and $\textbf{c}$ and obtain the final representation $\mathbf{kc} \in \mathbb{R}^{d_2}$. Essentially,  $\mathbf{kc}$ represents the CKG-aware representation learned from the utterances $x_{n-1}$ and $x_{n}$.
\subsubsection{Classification}
Finally, we concatenate the utterance representation $\mathbf{x} \in \mathbb{R}^{d_1}$ and the CKG-aware keyword and concept representation $\mathbf{kc} \in \mathbb{R}^{d_2}$, and then feed it into a linear transformation layer, followed by a softmax layer. The entire model is optimized by minimizing the negative log-likelihoods of all ground-truth next-turn keywords. 

\subsection{CKG-Guided Keyword Selection Strategy}
After obtaining a keyword distribution of the next utterance using the proposed next-turn keyword predictor, we propose a CKG-guided keyword selection strategy to select the most appropriate keyword for subsequent keyword-augmented response retrieval. Specifically, we select the keyword that is closer to the target than current keywords and has the highest probability. The distance between keywords is measured as the weighted path length between keywords on the CKG, computed by the Floyd-Warshall algorithm \cite{floyd1962algorithm}. Note that the edge weights on ConceptNet correlate positively with concept relatedness. Hence, we apply a reciprocal operation to the weights before computing path lengths. Essentially, our proposed strategy allows the agent to chat smoothly (by selecting the most likely next-turn keyword) while leading the conversation closer to the target keyword (by traversing to the target keyword via the most reasonable path on the CKG).

\subsection{Keyword-Augmented Response Retrieval}
The last module in our approach is a keyword-augmented response retrieval model, as illustrated in Figure \ref{fig: response retrieval}. At a high-level, it is a response retrieval model that selects the best candidate response given the context utterances and the predicted keywords.

\subsubsection{Utterance Representations}
The context utterance representation $\mathbf{X} \in \mathbb{R}^{N_x \times d}$ is obtained by the concatenation of two representations: 1) the flattened GRU encoded contextual representation and 2) the CKG-aware contextual concept representation, where $N_x$ denotes the total number of tokens and concepts in the context and $d$ denotes the hidden size of GRU and GGNN. Similarly, the candidate utterance representation $\mathbf{Y} \in \mathbb{R}^{N_y \times d}$ is obtained by: 1) the GRU encoded candidate representation and 2) the CKG-aware candidate concept representation, where $N_y$ denotes the total number of tokens and concepts in the candidate.

\subsubsection{Keyword Representations}
Besides utterance-based matching, we learn keyword-based matching to allow keyword-augmented response retrieval. To this end, we aim to select the candidate that best matches the predicted next-turn keywords given contextual utterances. Specifically, we first obtain the top predicted next-turn keywords using a trained next-turn keyword predictor. We then obtain the CKG-aware predicted keyword representation $\mathbf{K}_x \in \mathbb{R}^{N_{k_x} \times d}$ and candidate keyword representation $\mathbf{K}_y \in \mathbb{R}^{N_{k_y} \times d}$ from GGNN, where $N_{k_x}$ and $N_{k_y}$ denotes the number of predicted keywords and candidate keywords, respectively. In practice, following \cite{tang2019target}, we set $N_{k_x} = 3$, allowing top-3 keywords to be matched with candidate keywords.

\subsection{Matching}
We compute the matching score $s_u \in \mathbb{R}$ between context utterance representation $\mathbf{X} \in \mathbb{R}^{N_x \times d}$ and candidate utterance representation $\mathbf{Y} \in \mathbb{R}^{N_y \times d}$ as follows:
\begin{equation}
\label{eqn: utterance matching}
    s_u = \textit{dot}(\textit{max}(\mathbf{X}), \textit{max}(\mathbf{Y}))
\end{equation}
where $\textit{max}$ denotes max pooling along the sequence dimension, and $\textit{dot}$ denotes dot product.

Similarly, the matching score $s_k \in \mathbb{R}$ between predicted keyword representation $\mathbf{K}_x \in \mathbb{R}^{N_{k_x} \times d}$ and candidate keyword representation $\mathbf{K}_y \in \mathbb{R}^{N_{k_y} \times d}$ is computed as follows:
\begin{equation}
\label{eqn: keyword matching}
    s_k = \textit{dot}(\textit{max}(\mathbf{K}_x), \textit{max}(\mathbf{K}_y))
\end{equation}

The final matching score $s \in \mathbb{R}$ is obtained as follows:
\begin{equation}
\label{eqn: final matching}
    s = s_u + \lambda_k s_k
\end{equation}
where $\lambda_k$ denotes a hyper-parameter controlling the weight for keyword scores. We optimize the entire response retrieval model by minimizing the negative log-likelihood of the ground-truth response among all candidates.

\section{Experimental Settings}
\label{sec: experiments}
In this section, we introduce the datasets, evaluation metrics, baselines and model settings.
\subsection{Dataset}
\label{sec: dataset}
\begin{table}[!t]
\small
\centering
\begin{tabular}{cccccc}
\hline
\textbf{Dataset} & \textbf{Split} & \textbf{\#Conv.} & \textbf{\#Utter.} & \textbf{\#Key.} & \textbf{Avg. \#Key.}\\
\hline
\multirow{3}{*}{ConvAI2} & Train & 8950 & 132601 & 2678 & 1.78\\
 & Valid & 485 & 7244 & 2069 & 1.79\\
 & Test & 500 & 7194 & 1571 & 1.50\\
\hline
\multirow{3}{*}{Reddit} & Train & 112693 & 461810 & 2931 & 2.27\\
& Valid & 6192 & 25899 & 2851 & 2.25\\
& Test & 5999 & 24108 & 2846 & 2.30\\
\hline
\end{tabular}
\caption{Dataset statistics. \#Key. denotes the number of unique keywords and Avg. \#Key. denotes the average number of keywords per utterance.}
\label{table: dataset}
\end{table}
We use the ConvAI2 dataset proposed in \cite{zhang2018personalizing,dinan2019second} and preprocessed in \cite{tang2019target} in our experiments. Conversations in ConvAI2 are open-domain and cover a broad range of topics. In addition, we collect a large-scale open-domain conversation dataset from the social media Reddit\footnote{https://www.reddit.com/. We use the Pushshift dataset on Google BigQuery.}. The proposed Reddit dataset is collected from casual chats on the \textit{CasualConversation}\footnote{https://www.reddit.com/r/CasualConversation} and \textit{CasualUK}\footnote{https://www.reddit.com/r/CasualUK/} subreddits, where users chat freely with each other in any topic. Reddit is significantly larger and more diverse than ConvAI2.

Following \cite{tang2019target}, we use TF-IDF and part-of-speech (POS) features to extract keywords from both datasets. We use a maximum of 8 contextual utterances and each utterance is truncated to 30 tokens. The number of keywords for each utterance is capped at 10. We limit the vocabulary of both datasets to the most frequent 20K tokens.

In the task of next-turn keyword prediction, we remove keyword transitions not covered by our CKG, as discussed in Our Approach. In addition, we remove self-loops, i.e., a keyword transit to itself, in both training and evaluation examples to prevent the model from predicting keywords that exist in the context, because predicting self-loops would not lead the conversation to the target. After preprocessing, the average number of keyword candidates for ConvAI2 and Reddit are 158 and 201, respectively. The number of nodes/edges on CKG are 87K/221K and 97K/273K for ConvAI2 and Reddit, respectively. The statistics of the two datasets are presented in Table \ref{table: dataset}.
\begin{table*}[!t]
\small
\centering
\begin{tabular}{c|cccc|cccc}
\hline
&\multicolumn{4}{c|}{\textbf{ConvAI2}}&\multicolumn{4}{c}{\textbf{Reddit}}\\
\hline
\textbf{Model} & \textbf{R@1} & \textbf{R@3} & \textbf{R@5} & \textbf{P@1} & \textbf{R@1} & \textbf{R@3} & \textbf{R@5} & \textbf{P@1}\\
\hline
Random & 1.03$\pm$0.09 & 2.99$\pm$0.12 & 4.83$\pm$0.04 & 1.18$\pm$0.12 & 0.60$\pm$0.06 & 1.88$\pm$0.24 & 3.35$\pm$0.34 & 0.69$\pm$0.04\\

PMI & 16.96 & 34.15 & 46.39 & 19.11 & 6.90 & 16.06 & 22.98 & 7.79\\

Neural & 17.81$\pm$0.35 & 34.59$\pm$0.42 & 44.88$\pm$0.66 & 19.91$\pm$0.57 & 7.22$\pm$0.26 & 16.81$\pm$0.20 & 23.89$\pm$0.21 & 8.12$\pm$0.35\\

Kernel & 16.23$\pm$0.50 & 32.07$\pm$0.84 & 42.62$\pm$0.76 & 17.57$\pm$0.87 & 7.38$\pm$0.17 & 17.10$\pm$0.28 & 24.81$\pm$0.70 & 8.24$\pm$0.22\\

DKRN & 18.03$\pm$0.15 & 34.60$\pm$0.56 & 45.06$\pm$0.95 & 20.09$\pm$0.38 & 7.11$\pm$0.21 & 16.47$\pm$0.72 & 23.42$\pm$0.98 & 8.08$\pm$0.29\\
\hline

Ours (CKC) & \textbf{19.31}$\pm$0.44 & \textbf{36.26}$\pm$0.45 & \textbf{46.32}$\pm$0.57 & \textbf{21.98}$\pm$0.66 & \textbf{8.23}$\pm$0.31 & \textbf{17.83}$\pm$0.25 & \textbf{24.89}$\pm$0.12 & \textbf{9.17}$\pm$0.28\\
\hline
\end{tabular}
\caption{Test results (in \%) for next-turn keyword prediction. Results are averaged over 3 random seeds.}
\label{table: next-turn keyword prediction}
\end{table*}
\begin{table*}[!t]
\small
\centering
\begin{tabular}{c|cccc|cccc}
\hline
&\multicolumn{4}{c|}{\textbf{ConvAI2}}&\multicolumn{4}{c}{\textbf{Reddit}}\\
\hline
\textbf{Model} & \textbf{R@1} & \textbf{R@3} & \textbf{R@5} & \textbf{MRR} & \textbf{R@1} & \textbf{R@3} & \textbf{R@5} & \textbf{MRR}\\
\hline
PMI & 48.67$\pm$0.25 & 75.88$\pm$0.49 & 86.38$\pm$0.15 & 64.74$\pm$0.26 & 45.31$\pm$0.70 & 68.93$\pm$0.37 & 79.75$\pm$0.46 & 60.42$\pm$0.50\\

Neural & 47.93$\pm$0.47 & 75.53$\pm$0.62 & 86.36$\pm$0.20 & 64.25$\pm$0.38 & 44.96$\pm$0.21 & 68.75$\pm$0.27 & 79.59$\pm$0.23 & 60.18$\pm$0.22\\

Kernel & 48.55$\pm$0.51 & 75.57$\pm$0.32 & 86.04$\pm$0.04 & 64.47$\pm$0.37 & 44.55$\pm$0.33 & 68.47$\pm$0.24 & 79.66$\pm$0.38 & 59.92$\pm$0.30\\

DKRN & 48.44$\pm$0.34 & 75.78$\pm$0.20 & 86.83$\pm$0.16 & 64.64$\pm$0.17 & 44.92$\pm$0.45 & 68.84$\pm$0.45 & 79.59$\pm$0.65 & 60.19$\pm$0.44\\
\hline

Ours (CKC) & \textbf{59.90}$\pm$0.41 & \textbf{83.03}$\pm$0.31 & \textbf{92.15}$\pm$0.17 & \textbf{73.50}$\pm$0.26 & \textbf{50.02}$\pm$0.41 & \textbf{72.94}$\pm$0.33 & \textbf{82.87}$\pm$0.22 & \textbf{64.33}$\pm$0.35\\
\hline
\end{tabular}
\caption{Test results (in \%) for keyword-augmented response retrieval. Results are averaged over 3 random seeds.}
\label{table: keyword-augmented response retrieval}
\end{table*}
\subsection{Evaluation Metrics}
\label{sec: evaluation metrics}
\subsubsection{Turn-Level Evaluation} 
Following \cite{tang2019target, qin2020dynamic}, we use \textbf{R@k}, the recall at position k (=1, 3, 5) over all neighboring keywords, and \textbf{P@1}, the precision at the first position, for next-turn keyword prediction. Note that we have a smaller set of candidate keywords than that in \cite{tang2019target} because we only keep neighboring keywords as candidates. 

We use \textbf{R@k}, the recall at position k (=1, 3, 5) over all 20 candidate responses (a ground-truth response and 19 negative candidates), and \textbf{MRR}, the mean reciprocal rank, for keyword-augmented response retrieval.

\subsubsection{Dialogue-Level Evaluation}
Following \cite{tang2019target}, we measure the target success rate (\textbf{Succ.}) and number of turns (\textbf{\#Turns}) to reach the target for keyword-guided conversation evaluation using self-play simulations. We run self-play simulations for 1K conversations between each model and a base response retrieval model\footnote{This model respond passively to messages, which is the same as the base model used in \cite{tang2019target}.}. In addition, we measure target success rate (\textbf{Succ.}) and conversation smoothness (\textbf{Smo.}) using human evaluations with three annotators on 100 conversations for each model. The smoothness is rated in the [1, 5] scale, higher is better.

\subsection{Baselines and Model Settings}
\label{sec: baselines}
We compare our model with the following baselines: PMI \cite{tang2019target}, Neural \cite{tang2019target}, Kernel \cite{tang2019target} and DKRN \cite{qin2020dynamic}. We follow their released implementations\footnote{We fixed a bug in DKRN where the keyword transition mask is obtained using train+valid+test datasets.}. All baselines are trained and evaluated using the same filtered datasets as our model.

We initialize the embedding layer of all models using GloVe embedding of size 200 \cite{pennington2014glove}. All hidden sizes in GRU and GGNN are set to 200. We use one layer in GGNN and set $\lambda_k = 0.01$. We optimize our model using Adam \cite{kingma2014adam} with batch size of 32, an initial learning rate of 0.001 and a decay rate of 0.9 for every epoch.

\section{Result Analysis}
\label{sec: result analysis}
In this section, we present the experimental results, model analysis, case study and limitations.

\subsection{Next-Turn Keyword Prediction}
\label{sec: next-turn keyword prediction}

The results for next-turn keyword prediction are presented in Table \ref{table: next-turn keyword prediction}. Among all baselines except Random, the non-parameterized PMI performs worst, and Neural, Kernel and DKRN performs comparably on both datasets. Our proposed model achieves consistent better performance than all baselines across all metrics and datasets, suggesting that incorporating CKG triplets into keyword prediction helps. 
\begin{table}[!t]
\small
\centering
\begin{tabular}{c|cc|cc}
\hline
&\multicolumn{2}{c|}{\textbf{ConvAI2}}&\multicolumn{2}{c}{\textbf{Reddit}}\\
\hline
\textbf{Model} & \textbf{Succ. (\%)} & \textbf{\#Turns} & \textbf{Succ. (\%)} & \textbf{\#Turns}\\
\hline
PMI & 14.6 & 5.83 & 5.1 & 4.88\\
Neural & 18.9 & 6.07 & 11.1 & 5.99\\
Kernel & 20.7 & 5.89 & 10.6 & 5.83\\
DKRN & 25.6 & 4.54 & 18.4 & 4.42\\
\hline
Ours (CKC) & \textbf{28.9} & \textbf{4.23} & \textbf{22.7} & \textbf{4.19}\\
\hline
\end{tabular}
\caption{Self-play simulation results.}
\label{table: self-play}
\end{table}
\begin{table}[!t]
\small
\centering
\begin{tabular}{c|cc|cc}
\hline
&\multicolumn{2}{c|}{\textbf{ConvAI2}}&\multicolumn{2}{c}{\textbf{Reddit}}\\
\hline
\textbf{Model} & \textbf{Succ. (\%)} & \textbf{Smo.} & \textbf{Succ. (\%)} & \textbf{Smo.}\\
\hline
PMI & 16.0 & 3.05 & 6.3 & 2.68\\
Neural & 17.3 & 2.77 & 11.0 & 2.85\\
Kernel & 22.3 & 2.88 & 12.3 & 2.57\\
DKRN & 25.0 & 3.01 & 17.7 & 2.81\\
\hline
Ours (CKC) & \textbf{29.3} & \textbf{3.27} & \textbf{22.3} & \textbf{3.08}\\
\hline
\end{tabular}
\caption{Human evaluation results. Smo. denotes conversation smoothness.}
\label{table: human evaluation}
\end{table}
\subsection{Keyword-Augmented Response Retrieval}
\label{sec: keyword-augmented response retrieval}
\begin{table}[!t]
\small
\centering
\begin{tabular}{c|cc}
\hline
\multicolumn{3}{c}{\textbf{Next-Turn Keyword Prediction}}\\
\hline
\textbf{Model}&\multicolumn{2}{c}{\textbf{R@1}}\\
Ours (CKC)&\multicolumn{2}{c}{\textbf{19.31}$\pm$0.44}\\
- concepts&\multicolumn{2}{c}{18.56$\pm$0.31}\\
\hline
\multicolumn{3}{c}{\textbf{Keyword-Augmented Response Retrieval}}\\
\hline
\textbf{Model}&\multicolumn{2}{c}{\textbf{R@1}}\\
Ours (CKC)&\multicolumn{2}{c}{\textbf{59.90}$\pm$0.41}\\
- concepts&\multicolumn{2}{c}{53.11$\pm$0.43}\\
- keywords&\multicolumn{2}{c}{52.30$\pm$0.54}\\
\hline
\multicolumn{3}{c}{\textbf{Self-Play Simulation}}\\
\hline
\textbf{Model}&\textbf{Succ.} (\%)&\textbf{\#Turns}\\
Ours (CKC)&\textbf{28.9}&\textbf{4.23}\\
- CKG-based strategy&22.3&4.42\\
\hline
\end{tabular}
\caption{Ablation study (in \%) on ConvAI2.}
\label{table: ablation study}
\end{table}
\begin{table}[!t]
\small
\centering
\begin{tabular}{l}
\hline
Target: \textbf{music}\\
\hline
\textit{A}: Hey, how are you doing?\\
\textit{H}: I'm well, thanks. Working on a \textbf{party} I'm planning.\\
\textit{A}: I am sitting here listening to \textbf{pearl} \textbf{jam}, my favorite \textbf{band}.\\
\textit{H}: Super cool! Do you \textbf{sing}? I was just \textbf{singing} in my \textbf{shower}.\\
\textit{A}: No, but I was in \textbf{jazz} \textbf{band} in hs.\\
\textit{H}: Congrats! I love \textbf{music} and playing my \textbf{guitar} and \textbf{violin}.\\
\textit{A}: That's awesome! However, my favorite is \textbf{country} \textbf{music}.\\
\hline
\end{tabular}
\caption{Case study from self-play simulations on ConvAI2. \textit{A} denotes our model and \textit{H} denotes the base model.}
\label{table: case study}
\end{table}
The results for keyword-augmented response retrieval are presented in Table \ref{table: keyword-augmented response retrieval}. The baselines differ in which next-turn keyword prediction model is used. It is surprising that all baselines perform comparably regardless of the next-turn keyword prediction model. This may suggest that the baselines are unable to effectively leverage the predicted keyword information into response retrieval. Our model achieves substantially better performance than all baselines on both datasets. The performance improvement can be primarily attributed to 1) we additionally incorporate utterance-related CKG triplets into utterance representation learning; and 2) we propose an additional keyword matching module to match the predicted keywords with candidate keywords, whereas baselines directly match predicted keywords with candidate utterances.

\subsection{Keyword-Guided Conversation}
\label{sec: keyword-guided conversation}
The self-play simulation results for keyword-guided conversation are presented in Table \ref{table: self-play}. DKRN performs best among all baselines, which can be primarily attributed to its strategy of selecting keyword-related responses. This strategy requires a pool of confident candidates to select from. A larger pool will lead to higher success rate but lower smoothness because potentially less likely candidates can be selected. In all experiments, we set the pool size to 100. Our model also leverages this strategy but instead use weighted path lengths to measure keyword relatedness. Our model outperforms all baselines in both metrics on both datasets. Note that the success rates on ConvAI2 are consistently larger than that on Reddit across all models, which can be partially due to the higher next-turn keyword prediction accuracy on ConvAI2.

The human evaluation results are presented in Table \ref{table: human evaluation}. The results for success rate are similar to that in self-play simulations. Among all baselines, DKRN has slightly more robust performance in smoothness on both datasets. Our model obtains consistently better performance in both success rate and smoothness on both datasets, suggesting that our model can select confident candidates that are also related to the target keyword.

\subsection{Model Analysis}
\label{sec: model analysis}

Table \ref{table: ablation study} presents the ablation study of our model across multiple tasks on the ConvAI2 test set. In both next-turn keyword prediction and keyword-augmented response retrieval, removing concepts representation from our model leads to degraded performance in R@1, suggesting that CKG triplets are helpful in learning the semantic representation of utterances. In keyword-augmented response retrieval, unlike other baselines that do not leverage keywords effectively, our model performs noticeably worse when keywords are removed, showing that our design of matching keywords separately indeed contribute to the overall matching. Finally, we examine the impact of our CKG-guided keyword selection strategy on self-play simulations. The results in Table \ref{table: ablation study} show that replacing our CKG-based strategy by the embedding-based strategy \cite{tang2019target, qin2020dynamic} leads to worse performance in both success rate and number of turns.

\subsection{Case Study}
\label{sec: case study}
We present a case study from our self-play simulations in Table \ref{table: case study}. Our model can lead the conversation from a starting keyword ``party'' to the target keyword ``music" smoothly and fast.

\subsection{Limitations}
\label{sec: limitations}
One major limitation of existing approaches including ours is the mediocre accuracy of retrieving keyword-related responses (this is different from keyword-augmented response retrieval where the ground-truth responses do not necessarily correlate with the input keywords), which bottlenecks the overall target success rate. In fact, for both DKRN and our model, the target keyword can be successfully selected most of the time during self-play simulations, however, both models can not retrieve the keyword-related responses given the selected target keyword accurately. A potential solution to this problem is to train the keyword-augmented response retrieval model on datasets where input keywords and ground-truth responses are correlated, which is left to future work.

\section{Conclusion}
\label{sec: conclusion}
We study the problem of imposing conversational goals/keywords on open-domain conversational agents. The keyword transition module in existing approaches suffer from noisy datasets and unreliable transition strategy. In this paper, we propose to ground keyword transitions on commonsense and propose two GNN-based models for the tasks of next-turn keyword transition and keyword-augmented response retrieval, respectively. Extensive experiments show that our proposed model obtains substantially better performance on these two tasks than competitive baselines. In addition, the model analysis suggests that CKG triplets and our proposed CKG-guided keyword selection strategy are helpful in learning utterance representation and keyword transition, respectively. Finally, both self-play simulations and human evaluations show that our model can achieve better success rate, reach the target keyword faster, and produce smoother conversations than baselines.

\section*{Acknowledgments}
This research is supported, in part, by Alibaba Group through Alibaba Innovative Research (AIR) Program and Alibaba-NTU Singapore Joint Research Institute (JRI) (Alibaba-NTU-AIR2019B1), Nanyang Technological University, Singapore. This research is also supported, in part, by the National Research Foundation, Prime Minister's Office, Singapore under its AI Singapore Programme (AISG Award No: AISG-GC-2019-003) and under its NRF Investigatorship Programme (NRFI Award No. NRF-NRFI05-2019-0002). Any opinions, findings and conclusions or recommendations expressed in this material are those of the authors and do not reflect the views of National Research Foundation, Singapore. This research is also supported, in part, by the Singapore Ministry of Health under its National Innovation Challenge on Active and Confident Ageing (NIC Project No. MOH/NIC/COG04/2017 and MOH/NIC/HAIG03/2017).

\bibliography{aaai21}
\end{document}